\documentclass[10pt,twocolumn,letterpaper]{article}

\usepackage{wacv}              

\usepackage[accsupp]{axessibility}  

\definecolor{wacvblue}{rgb}{0.21,0.49,0.74}
\usepackage[pagebackref,breaklinks,colorlinks,allcolors=wacvblue]{hyperref}
\usepackage{float}
\usepackage{siunitx}
\usepackage{pifont}
\usepackage{multirow} 

\def\wacvPaperID{87} 
\def\confName{WACV}
\def\confYear{2026}

\title{Can Image Splicing and Copy-Move Forgery Be Detected by the Same Model? Forensim: An Attention-Based State-Space Approach}

\author{
Soumyaroop Nandi$^{1,2}$ \quad Prem Natarajan$^{1,2}$\\
$^{1}$USC Information Sciences Institute, Marina del Rey, CA, USA\\
$^{2}$USC Thomas Lord Department of Computer Science, Los Angeles, CA, USA\\
{\tt\small \{soumyarn, premkumn\}@usc.edu}
}

\begin{document}
\maketitle

\begin{abstract}
We introduce Forensim, an attention-based state-space framework for image forgery detection that jointly localizes both manipulated (target) and source regions. Unlike traditional approaches that rely solely on artifact cues to detect spliced or forged areas, Forensim is designed to capture duplication patterns crucial for understanding context. In scenarios like protest imagery, detecting only the forged region—e.g., a duplicated act of violence inserted into a peaceful crowd—can mislead interpretation, highlighting the need for joint source-target localization. Forensim outputs three-class masks (pristine, source, target) and supports detection of both splicing and copy-move forgeries within a unified architecture. We propose a visual state-space model that leverages normalized attention maps to identify internal similarities, paired with a region-based block-attention module to distinguish manipulated regions. This design enables end-to-end training and precise localization. Forensim achieves state-of-the-art performance on standard benchmarks. We also release CMFD\_Anything, a new dataset addressing limitations of existing copy-move forgery datasets. \noindent \href{https://github.com/SoumyaroopNandi/Forensim}{Project page and code}.
\end{abstract}   
\vspace{-5px}
\section{Introduction}
\label{sec:intro}


\begin{figure}[t]
    \centering
    \includegraphics[width=\columnwidth, height=12cm, keepaspectratio]{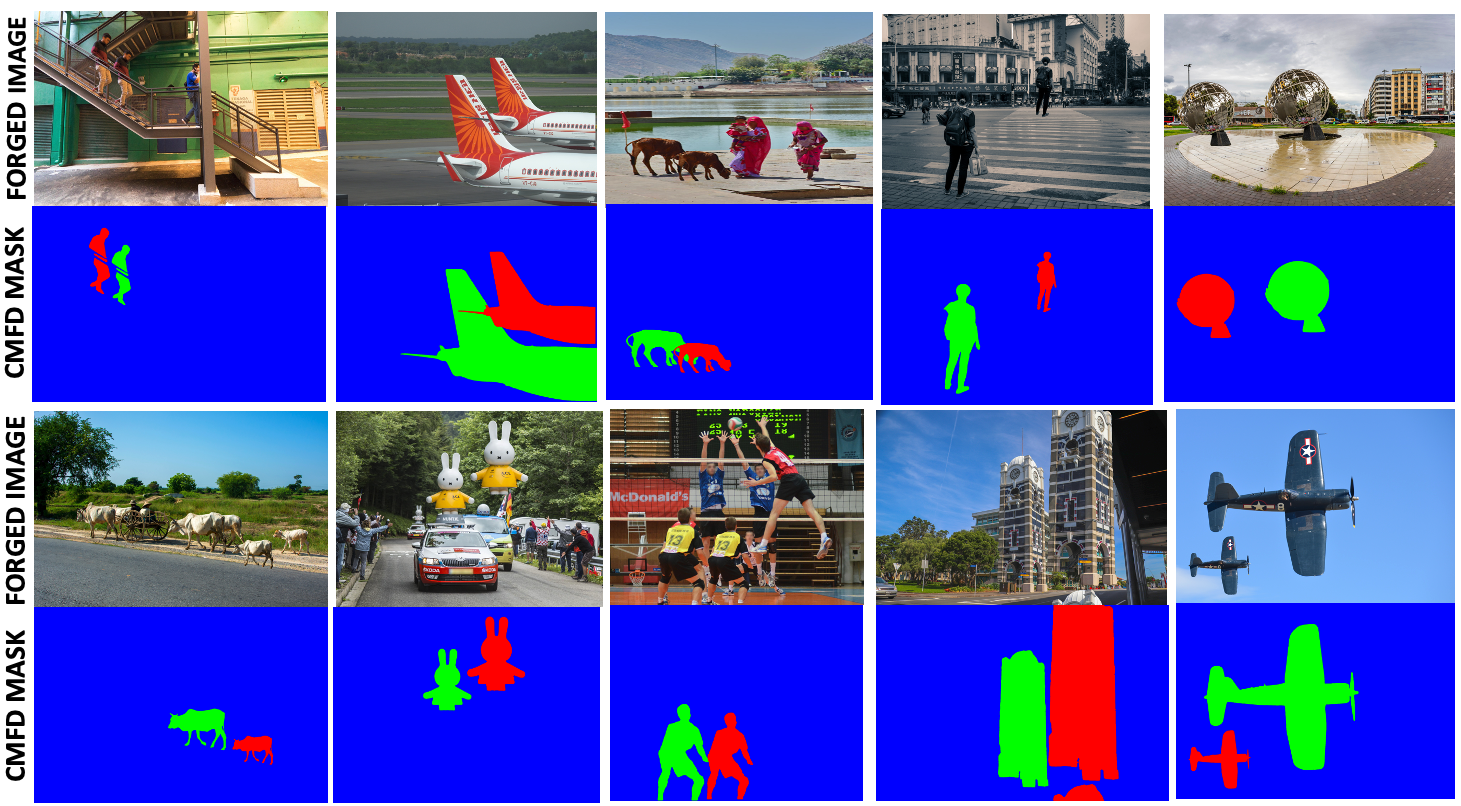}
    \caption{
        Proposed \texttt{CMFD\_Anything} samples-
        Rows show: (a) forged image, (b) RGB mask, (c) forged image, (d) RGB mask. 
        Source-target mask encodes \textcolor{blue}{untampered}, \textcolor{green}{source}, \& \textcolor{red}{target} regions.
    }
    \label{fig:cmfd_anything}
\end{figure}

\hspace*{1.2em} The proliferation of generative models has broadened image manipulation far beyond traditional editing tools. The challenge for today's researchers is to detect manipulated images, irrespective of the source of manipulation, be it generative models \cite{goodfellow2014generative,zhu2017unpaired,mirza2014conditional,pu2016variational,kingma2013auto}, tampered with by photo editing tools \cite{dhamo2020semantic, park2020swapping, vinker2020deep} or online social media filters \cite{wu2022robust}. Deceitful attackers may spread misinformation in the form of internet rumors \cite{wu2022robust}, fake news \cite{huh2018fighting}, deep fakes \cite{sabir2019recurrent, sabir2022monet}, forged satellite images \cite{horvath2021manipulation} plagiarized academic publications \cite{sabir2021biofors, sabir2022monet}. The prediction of the source of manipulation in the tampered images would be a vital contribution to the image manipulation detection community. Without the knowledge of the source of manipulation, tracing a tampered image becomes a challenging task. 

Tracing the source of image manipulation is possible if a model is trained to detect all forms of image tampering, whether involving image content or not. Image content-based forgeries can originate externally, where patches from an unrelated image are inserted (\textit{splicing forgery}), or internally, where parts of the same image are duplicated and repositioned (\textit{copy-move forgery}). Some forgers attempt to remove objects or regions from an image and seamlessly replace them with background pixels (\textit{inpainting forgery}).

Beyond direct content manipulation, images can also be subtly altered without modifying their content through various forms of \textit{image enhancement}, such as quantization, noise addition, resampling, blurring, morphing, compression, and histogram manipulation. Additionally, image tampering can stem from generative model-based synthesis \cite{wang2020cnn, corvi2023detection} or transformations introduced by social media \cite{wu2022robust}.

However, prior research has predominantly approached forgery detection from the perspective of image splicing, where models are trained with binary masks to capture artifact cues near manipulated regions. These approaches often assume that techniques effective for splicing generalize well to all types of manipulation, including copy-move forgeries. Traditional splicing detection models localize only the manipulated regions, without accounting for the duplicated source regions. In contrast, our proposed model, Forensim, is trained using three-class masks—\textcolor{blue}{pristine}, \textcolor{green}{source}, and \textcolor{red}{target}—to handle both splicing and copy-move forgeries in a unified framework, as illustrated in \hyperref[fig:splicing_setups]{Fig.~\ref*{fig:splicing_setups}}.

For example, in a forensic investigation, establishing a crime often requires not only identifying the manipulated evidence—such as perpetrator's fingerprints—but also tracing its origin, such as matching the fingerprint in a database. Analogously, in image manipulation detection and localization (IMDL), detecting a tampered region without locating its source can be insufficient, particularly when both exist within the same image. This observation motivates our approach, which reformulates the image forgery detection task as a three-class mask segmentation problem. Our framework is capable of distinguishing all relevant regions, and experimental results demonstrate that this design generalizes effectively across both image splicing/inpainting scenarios (binary masks) and copy-move forgery detection (CMFD), offering improved interpretability and robustness.

In this paper, we introduce a novel state-space model-based attention mechanism designed to accurately localize both the source and forged regions in IMDL. Specifically, we propose a similarity (Sim\_Attn) and manipulation (MSSA) attention modules, both of which leverage image affinity matrices to guide attention-based learning. Sim\_Attn identifies and highlights similar regions within an image, while MSSA distinguishes forged regions from naturally occurring identical patterns. By jointly learning these complementary attention mechanisms, our approach effectively mitigates misdetections and false alarms, addressing the key limitations of existing IMDL models. 

The attention modules efficiently capture long-range dependencies while maintaining linear computational complexity and low memory usage. Additionally, we normalize the attention weights in the discretization step of the Mamba formulation in \hyperref[eqn:2]{Eq.\ref*{eqn:2}}. This normalization ensures a well-balanced distribution of attention, enhancing robustness—particularly in fine-grained tasks such as CMFD. Furthermore, we have conducted extensive experiments with different MSSA blocks in \hyperref[fig:mssa_attn]{Fig.\ref*{fig:mssa_attn}} to develop a robust and effective CMFD architecture.

Training deep learning models for CMFD requires synthetic datasets, as open-source CMFD datasets are scarce. To address this, we introduce CMFD\_Anything, a benchmark dataset derived from high-quality Segment\_Anything images~\cite{kirillov2023segment}. Existing CMFD datasets, often created from MSCOCO~\cite{lin2014microsoft}, are limited and contain forgeries that are easily detectable by the naked eye. In contrast, CMFD\_Anything generates realistic, high-quality forgeries, enabling the development of robust models like Forensim to detect highly post-processed tampered images in \hyperref[fig:cmfd_anything]{Fig.\ref*{fig:cmfd_anything}}.


To unify splicing and CMFD, Forensim is trained on both spliced images with binary masks and CMFD images with source-target masks, enabling generalization across manipulation types. Our key contributions are:

\begin{enumerate}
\item[$\bullet$] We reformulate IMDL as a three-class training task, enabling unified detection of both CMFD and splicing forgeries—unlike prior work limited to binary-mask training focused on artifact-based splicing detection.
\item[$\bullet$] We propose similarity and manipulation attention modules within a state-space framework that effectively capture long-range dependencies with linear complexity.
\item[$\bullet$] We introduce \texttt{CMFD\_Anything}, a high-resolution copy-move forgery dataset derived from Segment Anything images, and demonstrate Forensim’s state-of-the-art performance on multiple CMFD benchmarks.
\end{enumerate}

\section{Related Work}
\label{sec:related_work}

\subsection{\noindent\textbf{Earlier IMDL Methods}} \label{early_IMDL}
\vspace{-5px}

Handcrafted feature-based methods for Image Manipulation Detection and Localization (IMDL) include ELA \cite{krawetz2007picture} (detects compression error differences between forged and pristine regions), NOI1 \cite{mahdian2009using} (models local noise using high-pass wavelet coefficients), and CFA1 \cite{ferrara2012image} (approximates camera filter array patterns). These methods typically focus on splicing detection. Early DNN-based studies also targeted specific forgeries, such as splicing \cite{cozzolino2015splicebuster}, copy-move \cite{cozzolino2015efficient}, removal \cite{zhu2018deep}, and enhancement \cite{bayar2016deep}. However, real-world manipulated images lack prior information about the forgery type, motivating the development of general forgery detection algorithms that can identify any manipulation.

\subsection{\noindent\textbf{Image Splicing Models and Training Datasets}} \label{image_splice_methods}

Most image splicing frameworks consist of an input feature extraction network, feature fusion (if multiple artifacts), and an anomaly detection network that classifies each pixel as tampered or pristine to generate a binary localization map. These networks are typically pre-trained on synthetic datasets, like MS-COCO \cite{lin2014microsoft}. ManTraNet \cite{wu2019mantra} detects 385 manipulation types using a bipartite end-to-end network but struggles with double JPEG compression artifacts. MVSS-Net \cite{chen2021image} uses noise distribution for feature generalization but fails to balance sensitivity and specificity. CAT-Net \cite{kwon2021cat} focuses on compression artifacts and is designed for image splicing only. IF-OSN \cite{wu2022robust} addresses artifacts from online social networks but struggles with traditional IMDL models. Trufor \cite{guillaro2023trufor} generates manipulation and confidence maps but does not predict manipulation sources. HiFi-Net \cite{guo2023hierarchical} treats IMDL as a multi-level classification task but fails when similar copy-move regions are present. None of these methods predict the source of manipulation, struggling in scenarios with similar copy-move regions.

\subsection{\noindent\textbf{CMFD Models and Training Datasets}} \label{cmfd_methods}

BusterNet~\cite{wu2018busternet} is the first end-to-end deep neural network for CMFD, featuring dual branches for source-target localization and trained on the synthetic USC-ISI CMFD dataset derived from MSCOCO~\cite{lin2014microsoft} and SUN2012~\cite{xiao2010sun}. DOA-GAN~\cite{islam2020doa} similarly adopts a GAN-based framework trained on the same dataset. However, both models struggle with complex copy-move forgeries and fail to generalize to image splicing, exhibiting high false alarm and misdetection rates, as shown in \hyperref[fig:qualitative_eval]{Fig.~\ref*{fig:qualitative_eval}}. The limited diversity, low complexity, and lack of availability of the USC-ISI dataset motivated us to develop the \texttt{CMFD\_Anything} dataset.

\subsection{\noindent\textbf{Structured State Space Models}} \label{state_space_models}
\vspace{-5px}

Structured state space models (SSM) offer an alternative to Transformer and CNN-based models, modeling long-range dependencies with linear scaling in sequence length. Recent advancements, such as S4~\cite{gu2021efficiently} and Mamba~\cite{gu2023mamba}, utilize structured SSMs for efficient long-range dependency modeling. VisionMamba~\cite{zhu2024vision} and VMamba~\cite{liu2024vmamba} further extend this for computer vision tasks. Despite their success, these models often lack task-specific design considerations, especially in IMDL. Our approach leverages a specialized SSM design optimized for similarity and manipulation detection, achieving better accuracy with lower complexity.

\begin{figure}[t]
    \centering
    \includegraphics[width=\columnwidth]{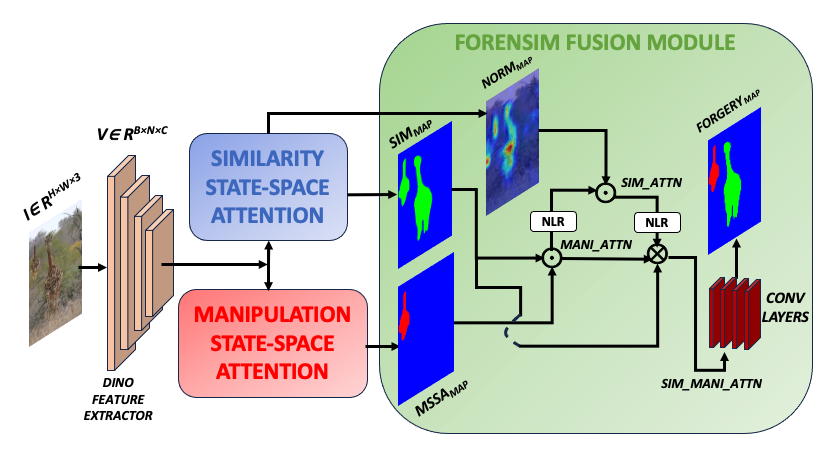} 
    \caption{Forensim Overview: Sim-Mani Attention and Fusion}
    \label{fig:forensim_fusion}
\end{figure}

\section{Proposed Method}

\subsection{State Space Model Overview}
\hspace*{1.5em} State Space Models (SSMs)~\cite{gu2021efficiently, gu2023mamba, liu2024vmamba} are linear time-invariant (LTI) systems that effectively capture complex sequential dependencies, similar to the Kalman Filter~\cite{kalman1960new}, which is designed for linear systems with simple dynamics. Given a continuous input sequence \( x(t) \in \mathbb{R} \), the system produces a continuous output \( y(t) \in \mathbb{R} \) through a hidden state \( h(t) \in \mathbb{R}^{N} \). The system's evolution is governed by the parameter \( \mathbf{A} \in \mathbb{R}^{N \times N} \), while \( \mathbf{B} \in \mathbb{R}^{N \times 1} \) and \( \mathbf{C} \in \mathbb{R}^{1 \times N} \) act as projection parameters. The continuous system follows these governing equations:

\vspace{-5pt}
\[
h'(t) = \textbf{A}h(t) + \textbf{B}x(t) 
\]
\vspace{-15pt}
\[
y(t) = \textbf{C}h(t)
\]

\vspace{-5px}
\hspace*{-1.5em}A timescale parameter \( \Delta \) is used to transform the continuous parameters \( \mathbf{A}, \mathbf{B} \) into their discrete counterparts, denoted as \( \mathbf{\bar{\textbf{A}}}, \mathbf{\bar{\textbf{B}}} \), as described in~\cite{gu2021efficiently, gu2023mamba}. A widely adopted approach for this transformation is the zero-order hold (ZOH)~\cite{gu2021efficiently}, which is defined as follows:

\vspace{-15pt}

\[
\bar{\textbf{A}} = \exp(\Delta {\textbf{A}}), \quad {\bar{\textbf{B}}} = (\Delta {\textbf{A}})^{-1} \left( \exp({\textbf{A}}) - \textbf{I} \right) \Delta {\textbf{B}}, \quad \bar{\textbf{C}} = {\textbf{C}},
\]

\vspace{-15pt}
\begin{equation}
\begin{aligned}
y_k &= \bar{\textbf{C}} h_k + \bar{\textbf{D}} x_k, \\
h_k &= \bar{\textbf{A}} h_{k-1} + \bar{\textbf{B}} x_k
\end{aligned}
\label{eqn:1}
\end{equation}


\noindent
\(\bar{\textbf{D}}\) works as a residual connection and \(\bar{\textbf{B}}\) can be approximated using the first-order Taylor series as:
\[
\bar{\textbf{B}} = \left( \exp(\textbf{A}) - \textbf{I} \right) \textbf{A}^{-1} \textbf{B} \approx (\Delta \textbf{A})(\Delta \textbf{A})^{-1} \Delta \textbf{B} = \Delta \textbf{B}
\]


\hspace*{-1.5em}\textbf{Selective Scan:} Mamba~\cite{gu2023mamba} excels at capturing and modeling complex interactions in long sequences through a selective scanning mechanism. The matrices \( \mathbf{B} \in \mathbb{R}^{L \times N} \), \( \mathbf{C} \in \mathbb{R}^{L \times N} \), and \( \Delta \in \mathbb{R}^{L \times D} \) are computed from the input \( \mathbf{x} \in \mathbb{R}^{L \times D} \), enabling the model to preserve contextual awareness. While this approach removes invariant parameters in one-dimensional temporal inputs, such as text, the 2D-Selective scan in VMamba~\cite{liu2024vmamba} extends this capability to two-dimensional images by incorporating spatial information and non-sequential structures through the Visual State Space Block (VSSM). 

In the first step, Cross-Scan, VSSM flattens input patches into four sequences, each following a distinct state space trajectory. Each sequence is processed independently and in parallel by separate VSSM instances to effectively capture long-range dependencies, as described in (Eqs.~\ref{eqn:1}, \ref{eqn:2}). Finally, in the Cross-Merge step, the processed sequences are reshaped and combined, generating an output map with global receptive fields in the two-dimensional space.

\subsection{Forensim Overview}

The architecture of the Forensim model is illustrated in \hyperref[fig:sim_attn]{Fig.\ref*{fig:forensim_fusion}}. It is specifically designed for three-class supervised forgery detection, employing state-space-based feature extraction and attention mechanisms to improve forgery localization. Given an input image \( I \in \mathbb{R}^{H \times W \times 3} \), the model utilizes the first four layers of an ImageNet-pretrained Vision Transformer backbone (DINO) to extract high-dimensional hierarchical feature representations \( V \in \mathbb{R}^{B \times N \times C} \), where \( B \) is the batch size, \( N = H \times W \), and \( C = 384 \) represents the embedding dimension for our proposed Similarity Attention Module. These features are subsequently flattened and processed through the proposed Similarity State Space Attention module, which constructs an affinity matrix \( Aff \in \mathbb{R}^{N \times N} \) by capturing self-similarity patterns within the image. This mechanism enhances the detection of duplicated regions, aiding in forgery identification.

\subsection{Similarity State Space Attention Module}
\label{sec:3.3}

Given an extracted feature map \( V \in \mathbb{R}^{B \times N \times C} \), the module computes an explicit affinity matrix \( Aff \in \mathbb{R}^{N \times N} \) that captures the similarity between different spatial locations, as shown in \hyperref[fig:sim_attn]{Fig.\ref*{fig:sim_attn}}. Leveraging the State Space Similarity mechanism described in \hyperref[eqn:1]{Eq.~\ref*{eqn:1}}, we normalize \( \bar{\textbf{C}} h_k \) such that:

\vspace{-5pt}
\begin{equation}
\begin{aligned}
y_k &= \bar{\textbf{C}} h_k / \bar{\textbf{C}} n_k + \bar{\textbf{D}} x_k, \\
h_k &= \bar{\textbf{A}} h_{k-1} + \bar{\textbf{B}} x_k
\end{aligned}
\label{eqn:2}
\end{equation}

\vspace{-5pt}
\noindent
Here, \( \mathbf{h}_k = \sum_{j=1}^{k} \mathbf{B}_j^\top x_j \quad \text{and} \quad \mathbf{n}_k = \sum_{j=1}^{k} \mathbf{B}_j \) \cite{gu2023mamba}. \hyperref[eqn:2]{Eq.\ref*{eqn:2}} defines the similarity attention with a global receptive field. Given a token \( k \) and its preceding token \( j \), where \( j \leq k \), each \( \bar{\textbf{A}} \) aggregates information from all \( \mathbf{x_k} \) and \( \mathbf{y_k} \). Similarity Attention divides the output by $\bar{\textbf{C}} n_k$, to ensure that the attention weights sum up to 1. 

We incorporate Rotary Positional Embeddings (RoPE)~\cite{su2024roformer} into the similarity attention mechanism to ensure that the affinity matrix captures spatial dependencies across the image effectively. Additionally, an ELU activation function is applied to \( \mathbf{V_k} \) prior to normalization to enhance stability during similarity computation. The Similarity Attention module computes an explicit affinity matrix via a dot product between the transformed representations \( \bar{\mathbf{C}} \) and \( \bar{\mathbf{B}} \), as defined in \hyperref[eqn:2]{Eq.~\ref*{eqn:2}}. Given an input \( V \in \mathbb{R}^{B \times N \times C} \), the vectors \( \bar{\mathbf{C}} \) and \( \bar{\mathbf{B}} \) are computed as:

\begin{figure}[t]
    \centering
    \includegraphics[width=1.1\columnwidth]{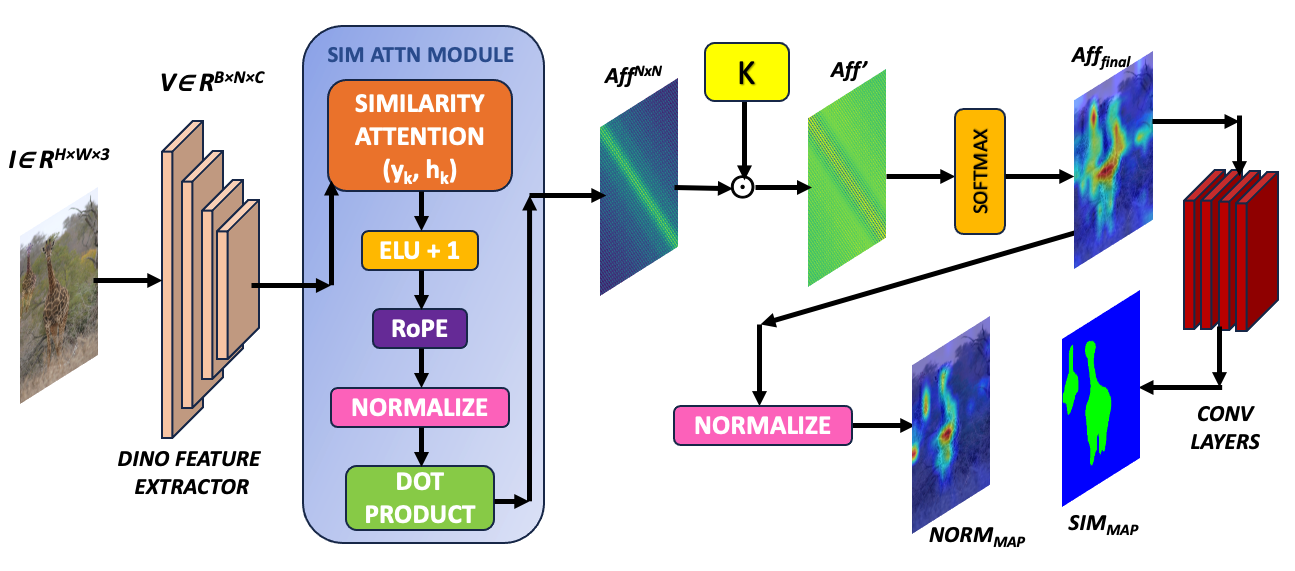} 
    \caption{Similarity State Space Attention Module}
    \label{fig:sim_attn}
\end{figure}

\vspace{-10pt}
\begin{equation}
\bar{\textbf{C}}, \bar{\textbf{B}} = \text{ELU}(\text{Linear}(V_{k})) + 1.0
\label{eqn:3}
\end{equation}

\noindent
These features are subsequently modulated using RoPE. Specifically, they are reshaped, augmented with rotary positional encoding, and then normalized as follows:

\vspace{-10pt}
\begin{equation}
\bar{\textbf{C}} = \frac{\text{RoPE}(\bar{\textbf{C}})}{\|\bar{\textbf{C}}\|_2}, \quad \bar{\textbf{B}} = \frac{\text{RoPE}(\bar{\textbf{B}})}{\|\bar{\textbf{B}}\|_2}
\label{eqn:4}
\end{equation}

\vspace{-5pt}
\noindent
Finally, the explicit affinity matrix is computed using the dot product between the modulated feature representations:

\vspace{-10pt}
\begin{equation}
Aff = \bar{\textbf{C}} \bar{\textbf{B}}^\top, \quad Aff \in \mathbb{R}^{B \times N \times N}
\label{eqn:5}
\end{equation}

\vspace{-5pt}
\noindent
This yields a structured affinity representation that captures spatial relationships and preserves computational efficiency.

When calculating the similarity attention of an image, the affinity matrix \( Aff \) exhibits higher values along the diagonal, as these values represent the correlation of a region with itself. To resolve this issue, we define an operation \( K \), as follows:

\vspace{-15pt}
\begin{equation}
K(p, q, p', q') = \frac{(p - p')^2 + (q - q')^2}{(p - p')^2 + (q - q')^2 + \sigma^2}
\label{eqn:6}
\end{equation}

\vspace{-5pt}
\noindent
We obtain a new affinity matrix \( Aff' = Aff \odot K \), where \( \odot \) denotes the element-wise dot product.

To refine the affinity matrix, a bidirectional softmax operation is applied, ensuring that the final similarity values are mutually reinforced between spatial locations by utilizing the patch-matching strategy from \cite{Cheng_2019_CVPR}, we compute the likelihood that a patch in the \(p\)-th row matches with a patch in the \(q\)-th column of \(Aff'\) to compute the final affinity matrix \(Aff_{final}\) as follows:

\vspace{-15pt}
\begin{equation}
Aff_{row}(p, q) = \frac{\exp(\alpha Aff'[p, q])}{\sum_{q' = 1}^{N} \exp(\alpha Aff'[p, q'])},
\label{eqn:7}
\end{equation}

\vspace{-15pt}
\begin{equation}
Aff_{col}(p, q) = \frac{\exp(\alpha Aff'[p, q])}{\sum_{p' = 1}^{N} \exp(\alpha Aff'[p', q])},
\label{eqn:8}
\end{equation}

\vspace{-15pt}
\begin{equation}
Aff_{final}(p, q) = Aff_{row}(p, q) \cdot Aff_{col}(p, q).
\label{eqn:9}
\end{equation}

\begin{figure}[t]
    \centering
    \includegraphics[width=0.8\columnwidth]{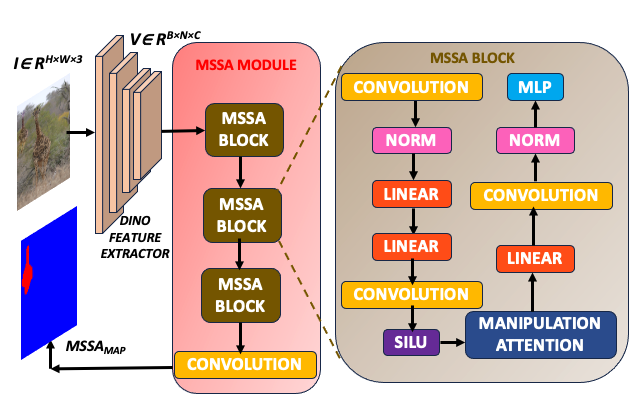} 
    \caption{Manipulation State Space Attention Module}
    \label{fig:mssa_attn}
\end{figure}

\vspace{-5pt}
\noindent
Here, \( \alpha \) is a hyperparameter, set to 5. The final affinity matrix \( Aff_{final} \in \mathbb{R}^{N \times N} \) is then refined using a Convolutional module, which comprises four convolutional layers followed by a SiLU activation. These layers process the affinity matrix to generate the similarity attention map \(Sim_{map} \in \mathbb{R}^{\sqrt{N} \times \sqrt{N}} \), which highlights the top-k most similar regions per pixel. 

We also obtain a normalized affinity map \(Norm_{map} \in \mathbb{R}^{N \times \sqrt{N} \times \sqrt{N}} \) from \hyperref[eqn:6]{Eq.\ref*{eqn:6}} to capture spatial dependencies across the entire image, such that:

\vspace{-15pt}
\begin{equation}
\text{Norm}_{\text{Map}}(p, q) = \frac{(Aff_{\text{final}}(p, q))}{\left( \sum_{q' = 1}^{N} Aff_{\text{final}}(p, q') \right)}
\label{eqn:10}
\end{equation}

\begin{figure}[t]
    \centering
    \begin{subfigure}[t]{0.49\columnwidth}
        \includegraphics[width=\linewidth]{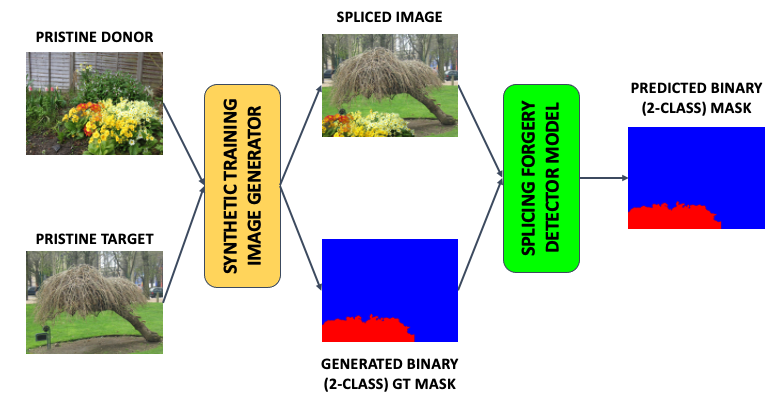}
        \caption*{(a)}
        \label{fig:splicing_Training_baseline}
    \end{subfigure} \hfill
    \begin{subfigure}[t]{0.5\columnwidth}
        \includegraphics[width=\linewidth]{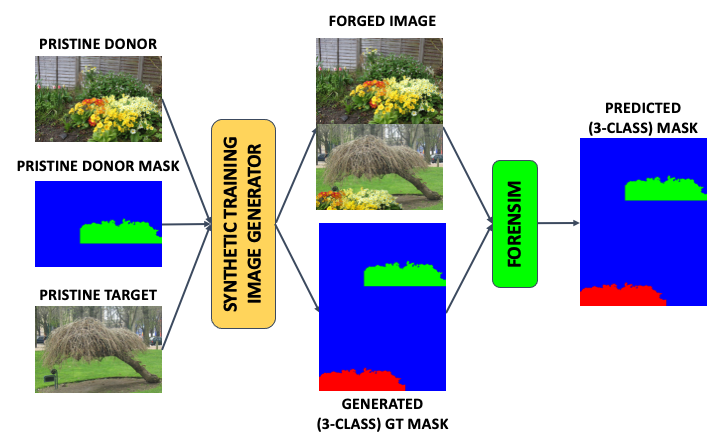}
        \caption*{(b)}
        \label{fig:splicing_Training_forensim}
    \end{subfigure}
    \caption{
        Splicing Training: (a) Baselines supervise only with \textcolor{red}{target} masks. 
        (b) Forensim uses both \textcolor{green}{source} and \textcolor{red}{target} masks.
    }
    \label{fig:splicing_setups}
\end{figure}

\subsection{Manipulation State Space Attention Module}
\label{sec:3.4}

The Multi-Level Manipulation State Space Attention (MSSA) module enhances tampered region detection by using a multi-scale state space attention mechanism composed of three \( MSSA_{Block} \) units. Each block processes the input with a specified number of attention heads to capture both global context and local feature details, addressing the computational challenges of traditional self-attention methods for high-resolution images.

Given an extracted feature \( V \in \mathbb{R}^{B \times N \times C} \), the three MSSA blocks capture multi-scale information, and their outputs are averaged and reshaped to the original spatial dimensions. A final convolution layer followed by SiLU activation produces the \( \text{MSSA}_{\text{map}} \), which highlights potential forgery regions while suppressing background noise, as illustrated in \hyperref[fig:mssa_attn]{Fig.~\ref*{fig:mssa_attn}}.

Manipulation Attention in \hyperref[fig:mssa_attn]{Fig.\ref*{fig:mssa_attn}} incorporates Locally Enhanced Positional Encoding (LePE) \cite{dong2022cswin} and Rotary Position Embedding (RoPE) \cite{su2024roformer} to enhance spatial awareness. Given an extracted input \( V \in \mathbb{R}^{B \times N \times C} \), the $\bar{\textbf{C}}$ and $\bar{\textbf{B}}$ from \hyperref[eqn:2]{Eq.\ref*{eqn:2}} are computed as: 

\vspace{-10pt}
\begin{equation}
\bar{\textbf{C}}, \bar{\textbf{B}} = \text{ELU}(\text{Linear}(V_{k})) + 1.0
\label{eqn:11}
\end{equation}

\vspace{-5pt}
\noindent
The representation is further modulated using RoPE, and attention is computed efficiently as:

\vspace{-15pt}
\begin{equation}
p = \frac{1}{\bar{\textbf{C}} \bar{\textbf{B}}^\top + \epsilon}, \quad \bar{\textbf{B}}V_{k} = (\bar{\textbf{B}}^\top N^{-0.5}) (V_{k} N^{-0.5})
\label{eqn:12}
\end{equation}

\vspace{-10pt}
\begin{equation}
V_{k} = (\bar{\textbf{C}} \cdot \text{RoPE}(\bar{\textbf{B}})) \bar{\textbf{B}}V_{k} \cdot p + \text{LePE}(V_{k})
\label{eqn:13}
\end{equation}

\noindent
Local Positional Encoding (LePE), implemented via depthwise convolution, enhances local feature aggregation and improves representational capacity. The generated \( V_k \) is passed through a combination of linear, CPE, normalization, and MLP layers to produce the MSSA Block output.

The \( \text{MSSA}_{\text{Block}} \) comprises a convolutional patch embedding (CPE) layer, normalization layers, and multi-head attention mechanisms. It sequentially applies convolution, normalization, and SiLU activation, followed by the manipulation state-space attention mechanism defined in \hyperref[eqn:2]{Eq.~\ref*{eqn:2}}, and a feed-forward network (MLP). Residual connections and drop path regularization further refine the output, enabling the MSSA Block to capture fine-grained spatial patterns for tampered region detection while maintaining computational efficiency.

\subsection{Forensim Fusion Module}

To efficiently integrate extracted features, we propose a Non-Local Refinement (NLR) module that propagates information across spatial locations using the similarity map \( \text{Sim}_{\text{map}} \) and normalization map \( \text{Norm}_{\text{map}} \) from Section~\ref{sec:3.3}, along with the manipulation self-structure attention map \( \text{MSSA}_{\text{map}} \) from Section~\ref{sec:3.4}. This operation enhances contextual coherence by reinforcing semantically meaningful regions and suppressing irrelevant background signals. We formalize the NLR operation as a non-local weighted aggregation of features:

\vspace{-15pt}
\begin{equation}
    \text{NLR}(F)_i = \sum_{j} A_{ij} \cdot F_j, \quad A_{ij} = \frac{\exp(S_{ij})}{\sum_k \exp(S_{ik})}
    \label{eqn:nlr}
\end{equation}

\vspace{-5pt}
\noindent
where \( F \in \mathbb{R}^{C \times N} \) is the input feature map, \( S_{ij} \) is the similarity between locations \( i \) and \( j \), and \( A_{ij} \) is the softmax-normalized attention weight. This operation allows each spatial feature to integrate context from semantically related regions, guided by similarity and manipulation cues.

We apply this refinement to both the manipulation and similarity attention pathways as follows:

\vspace{-5pt}
\begin{equation}
    \text{Mani\_Attn} = \text{NLR}(\text{MSSA}_{\text{map}} \odot \text{Sim}_{\text{map}})
    \label{eqn:16}
\end{equation}

\vspace{-10pt}
\begin{equation}
    \text{Sim\_Attn} = \text{NLR}(\text{Mani\_Attn} \odot \text{Norm}_{\text{map}})
    \label{eqn:14}
\end{equation}

where \( \odot \) denotes the Hadamard product.

We then construct the fused representation \( \text{Sim\_Mani\_Attn} \in \mathbb{R}^{C \times \sqrt{N} \times \sqrt{N}} \) by channel-wise concatenating the aggregated features, as illustrated in \hyperref[fig:forensim_fusion]{Fig.~\ref*{fig:forensim_fusion}}:

\vspace{-10pt}
\begin{equation}
    \text{Sim\_Mani\_Attn} = \text{Mani\_Attn} \otimes \text{Sim\_Attn} \otimes \text{Sim}_{\text{map}}
    \label{eqn:18}
\end{equation}

\vspace{-5pt}
\noindent
where $\otimes$ denotes channel-wise concatenation.

This fused representation captures global self-similarity, local manipulation context, and fine-grained structural cues. It is further processed by a sequence of convolutional layers with SiLU activations (as described in Section~\ref{sec:3.3}) to produce a pixel-level forgery mask and an image-level detection score. This structured design enables Forensim to generalize robustly across diverse manipulation types through context-aware, semantically aligned feature integration.

\begin{table}[t]
    \centering
    \scriptsize
    \caption{Image Quality Assessment (IQA) on CMFD datasets. $\uparrow$ indicates higher is better; $\downarrow$ indicates lower is better.}
    \vspace{-2mm}
    \setlength{\tabcolsep}{2pt}
    \begin{tabular}{l|c|c|c|c}
        \toprule
        \textbf{Method} & \textbf{Type} & \textbf{CASIA CMFD~\cite{dong2013casia}} & \textbf{CoMoFoD~\cite{tralic2013comofod}} & \textbf{CMFD\_Anything} \\
        \midrule
        PaQ-2-PiQ~\cite{ying2020patches} ($\uparrow$) & No-Ref & 3.64 & 3.42 & \textbf{4.11} \\
        CLIP-IQA~\cite{wang2023exploring} ($\uparrow$) & No-Ref & 6.58 & 6.42 & \textbf{7.23} \\
        \midrule
        SSIM~\cite{wang2004image} ($\uparrow$) & Full-Ref & 0.832 & 0.779 & \textbf{0.881} \\
        LPIPS~\cite{zhang2018unreasonable} ($\downarrow$) & Full-Ref & 0.179 & 0.243 & \textbf{0.124} \\
        \bottomrule
    \end{tabular}
    \vspace{-2mm}
    \label{tab:cmfd_quality_scores}
\end{table}

\section{Training and Implementation}
\vspace{-2px}

\begin{table*}
    \centering
    \scalebox{0.7}{
    \begin{tabular}{l|ccc|ccc|ccc|c|c|c|c}
        \toprule
        \textbf{Methods} & \multicolumn{3}{c}{\textbf{Precision (Localization)}} & \multicolumn{3}{c}{\textbf{Recall (Localization)}} & \multicolumn{3}{c}{\textbf{F1 (Localization)}} & \textbf{AUC} & \textbf{Precision} & \textbf{Recall} & \textbf{F1} \\
        \cmidrule(lr){2-4} \cmidrule(lr){5-7} \cmidrule(lr){8-10}
        & P & S & T & P & S & T & P & S & T & (Loc.) & (Detection) & (Detection) & (Detection) \\
        \midrule
        \multicolumn{14}{c}{\textbf{USC-ISI CMFD Test Set~\cite{wu2018busternet}}} \\
        \midrule
        BusterNet~\cite{wu2018busternet} & 93.71 & 55.85 & 53.84 & 99.01 & 38.26 & 48.73 & 96.15 & 40.84 & 48.33 & 0.64 & 89.26 & 80.14 & 84.45 \\
        ManTra-Net~\cite{wu2019mantra} & 93.50 & 8.66 & 48.53 & \underline{99.22} & 2.28 & 28.43 & 96.08 & 2.97 & 30.58 & 0.53 & 68.72 & 85.82 & 76.32 \\
        DOA-GAN~\cite{islam2020doa} & 96.99 & 76.30 & 85.60 & 98.87 & 63.57 & 80.45 & 97.69 & 66.58 & 81.72 & 0.83 & 96.83 & 96.14 & 96.48 \\
        HiFi-Net~\cite{guo2023hierarchical} & 92.80 & 7.10 & 46.00 & 98.80 & 1.90 & 26.00 & 95.30 & 2.50 & 29.00 & 0.52 & 66.00 & 84.00 & 74.00 \\
        TruFor~\cite{guillaro2023trufor} & 96.88 & 77.10 & 86.42 & 99.01 & 65.92 & 81.93 & 97.99 & 67.82 & 82.89 & 0.85 & 96.95 & 96.88 & 96.59 \\
        SparseViT~\cite{su2025can} & 97.01 & 77.85 & 87.23 & 99.10 & 66.91 & 82.47 & 98.06 & 68.29 & \underline{83.44} & 0.86 & 97.10 & 97.08 & 96.71 \\
        Forensim w/o Sim\_Attn & 92.62 & 69.43 & 78.67 & 97.11 & 57.64 & 72.56 & 96.23 & 61.36 & 74.69 & 0.78 & 86.95 & 81.67 & 80.06 \\
        Forensim w/o MSSA  & 91.48 & 68.87 & 77.39 & 96.81 & 58.35 & 71.43 & 95.75 & 60.84 & 72.91 & 0.77 & 85.94 & 80.62 & 79.48 \\
        Forensim w/o Detection & \underline{97.23} & \underline{78.43} & \underline{88.63} & 98.91 & \underline{68.92} & \underline{82.64} & \underline{98.11} & \underline{69.37} & 82.14 & \underline{0.88} & \underline{97.22} & \underline{97.38} & \underline{96.84} \\
        \textbf{Forensim}  & \textbf{97.43} & \textbf{79.61} & \textbf{89.29} & \textbf{99.34} & \textbf{70.49} & \textbf{99.67} & \textbf{99.92} & \textbf{74.83} & \textbf{84.48} & \textbf{0.90} & \textbf{98.47} & \textbf{97.64} & \textbf{96.98} \\
        \midrule
        \multicolumn{14}{c}{\textbf{CMFD Anything Test Set (Ours)}} \\
        \midrule
        BusterNet~\cite{wu2018busternet} & 47.34 & 36.88 & 35.16 & 53.42 & 26.78 & 34.96 & 47.87 & 28.34 & 31.26 & 0.47 & 44.56 & 42.61 & 43.12 \\
        ManTra-Net~\cite{wu2019mantra} & 48.65 & 7.16 & 37.48 & 64.98 & 2.21 & 24.85 & 48.16 & 2.41 & 27.89 & 0.44 & 52.62 & 64.93 & 57.14 \\
        DOA-GAN~\cite{islam2020doa} & 53.48 & 48.72 & 52.67 & 61.83 & 32.12 & 43.74 & 73.36 & 37.62 & 44.83 & 0.60 & 70.06 & 73.44 & 71.42 \\
        HiFi-Net~\cite{guo2023hierarchical} & 47.50 & 6.50 & 36.00 & 73.00 & 1.90 & 23.00 & 47.20 & 2.10 & 26.00 & 0.43 & 51.00 & 63.00 & 55.50 \\
        TruFor~\cite{guillaro2023trufor} & 56.91 & 50.83 & 57.63 & 65.42 & 36.21 & 47.32 & 77.83 & 41.29 & 49.51 & 0.63 & 72.45 & 74.62 & 73.01 \\
        SparseViT~\cite{su2025can} & 57.74 & 52.14 & 59.95 & 67.88 & 38.04 & 49.89 & 79.36 & 43.16 & 51.73 & 0.65 & 73.12 & 75.81 & 74.16 \\
        Forensim w/o Sim\_Attn & 52.68 & 47.39 & 50.64 & 58.19 & 30.87 & 43.26 & 70.83 & 35.87 & 41.27 & 0.58 & 67.58 & 68.92 & 66.84 \\
        Forensim w/o MSSA  & 51.67 & 47.18 & 49.21 & 56.82 & 31.58 & 42.96 & 68.12 & 34.59 & 40.26 & 0.57 & 66.89 & 67.14 & 65.72 \\
        Forensim w/o Detection & \underline{58.45} & \underline{53.72} & \underline{61.28} & \underline{68.76} & \underline{40.74} & \underline{51.96} & \underline{81.23} & \underline{48.73} & \underline{53.47} & \underline{0.68} & \underline{74.94} & \underline{76.83} & \underline{75.17} \\
        \textbf{Forensim}  & \textbf{59.41} & \textbf{54.82} & \textbf{64.77} & \textbf{69.74} & \textbf{41.91} & \textbf{53.43} & \textbf{82.67} & \textbf{50.61} & \textbf{54.16} & \textbf{0.70} & \textbf{75.12} & \textbf{77.42} & \textbf{75.68} \\
        \bottomrule
    \end{tabular}
    }
    \caption{CMFD results on USC-ISI CMFD~\cite{wu2018busternet} and CMFD-Anything (ours) test sets. Localization uses pixel-level precision, recall, F1 (Pristine(P)/Source(S)/Target(T) regions), and AUC. Detection uses image-level metrics. \textbf{Bold} = best per column, \underline{underline} = second best.}
    \label{tab:cmfd_combined}
\end{table*}


\subsection{Image Splicing using CMFD Training Setup} 
Forensim is designed to detect both splicing and copy-move forgeries using a unified architecture that outputs 3-class masks (\textcolor{blue}{pristine}, \textcolor{green}{source}, \textcolor{red}{target}) for all manipulation types, as shown in \hyperref[fig:qualitative_eval]{Fig.~\ref*{fig:qualitative_eval}}. Unlike prior splicing models~\cite{wu2019mantra, chen2021image, kwon2021cat, guillaro2023trufor, guo2023hierarchical, su2025can}, which are trained with binary masks \hyperref[fig:splicing_Training_baseline]{(Fig.~\ref*{fig:splicing_setups}a)}, capture only artifact cues, and do not model duplication or source regions, our proposed setup \hyperref[fig:splicing_Training_forensim]{(Fig.~\ref*{fig:splicing_setups}b)} uses 3-class supervision—even for spliced images—enabling the model to learn explicit source-target relationships. This joint modeling of Sim\_Map and MSSA\_Map allows the network to generalize across manipulation types without task-specific datasets or architectural modifications. While we do not claim to be the first to address CMFD, our key novelty lies in framing Forensim as the first general-purpose IMDL framework trained using a three-class-based training formulation, rather than targeting a specific task like splicing or copy-move. Importantly, we argue that three-class-based training is not a simplified subset of splicing; rather, localizing both source and target regions constitutes a more structured and fine-grained supervision signal than detecting tampering boundaries or artifact regions alone.

\begingroup
\captionsetup[table]{aboveskip=1pt,belowskip=0pt}
\captionsetup[figure]{aboveskip=1pt,belowskip=0pt}

\begin{figure*}[t]
    \centering
    \scriptsize
    \begin{minipage}[t]{0.46\textwidth}
        \captionsetup{justification=centering, singlelinecheck=false}
        \captionof{table}{Pixel and Image-level on CASIA and CoMoFoD CMFD datasets. \textbf{Bold} = \textbf{Best}, \underline{Underline} = \underline{Second-Best}.}
        \label{tab:casia_comofod_merged}
        \scalebox{0.86}{
        \begin{tabular}{l|c|c|c|c||c|c|c}
            \hline
            \textbf{Method} & \multicolumn{3}{c|}{\textbf{Localization}} & \textbf{AUC} & \multicolumn{3}{c}{\textbf{Detection}} \\
            \cline{2-4} \cline{6-8}
            & Prec. & Rec. & F1 & (Loc.) & Prec. & Rec. & F1 \\
            \hline
            \multicolumn{8}{c}{\textbf{CASIA CMFD \cite{dong2013casia}}} \\
            \hline
            BusterNet~\cite{wu2018busternet} & 42.15 & 30.54 & 33.72 & 0.38 & 48.34 & 75.12 & 58.82 \\
            ManTraNet~\cite{wu2019mantra} & 38.12 & 27.42 & 30.85 & 0.32 & 52.13 & 67.14 & 58.76 \\
            DOA-GAN~\cite{islam2020doa} & 54.70 & 39.67 & 41.44 & 0.46 & 63.39 & 77.00 & 69.53 \\
            TruFor~\cite{guillaro2023trufor} & \underline{55.67} & 41.83 & 43.62 & \underline{0.47} & \underline{78.92} & 86.32 & 82.43 \\
            SparseViT~\cite{su2025can} & 54.32 & \underline{42.91} & \underline{45.87} & 0.46 & 77.81 & \underline{87.43} & \underline{83.23} \\

            \textbf{Forensim} & \textbf{61.87} & \textbf{47.24} & \textbf{58.12} & \textbf{0.57} & \textbf{84.24} & \textbf{91.71} & \textbf{89.12} \\
            w/o Sim\_Attn & 54.78 & 41.42 & 42.91 & 0.46 & 77.42 & 85.03 & 80.92 \\
            w/o MSSA & 54.32 & 40.87 & 42.48 & 0.45 & 77.23 & 84.87 & 80.76 \\
            \hline
            \multicolumn{8}{c}{\textbf{CoMoFoD CMFD \cite{tralic2013comofod}}} \\
            \hline
            BusterNet~\cite{wu2018busternet} & 51.25 & 28.20 & 35.34 & 0.39 & 53.20 & 57.41 & 55.22 \\
            ManTraNet~\cite{wu2019mantra} & 36.11 & 25.48 & 28.34 & 0.30 & 50.34 & 54.92 & 52.48 \\
            DOA-GAN~\cite{islam2020doa} & 48.42 & 37.84 & 36.92 & 0.42 & 60.38 & 65.98 & 63.05 \\
            TruFor~\cite{guillaro2023trufor} & 49.83 & \underline{38.74} & 36.54 & \underline{0.43} & 66.91 & \underline{71.25} & 68.94 \\
            SparseViT~\cite{su2025can} & \underline{51.15} & 37.20 & \underline{37.92} & 0.42 & \underline{67.70} & 70.41 & \underline{69.83} \\

            \textbf{Forensim} & \textbf{56.41} & \textbf{43.57} & \textbf{47.82} & \textbf{0.51} & \textbf{76.58} & \textbf{77.94} & \textbf{75.82} \\
            w/o Sim\_Attn & 49.51 & 36.83 & 35.91 & 0.41 & 66.35 & 69.01 & 67.31 \\
            w/o MSSA & 48.89 & 36.20 & 35.47 & 0.40 & 66.02 & 68.74 & 67.10 \\
            \hline
        \end{tabular}
        }
    \end{minipage}
    \hfill
    \begin{minipage}[t]{0.53\textwidth}
        \centering
        \vspace{-4pt}
        \captionsetup{justification=centering}
        \captionof{figure}{Qualitative results on four CMFD datasets. Rows:Datasets; Columns:Models. Zoom in to view \textcolor{blue}{Untampered}, \textcolor{green}{Source}, \textcolor{red}{Target}.}
        \includegraphics[width=\linewidth]{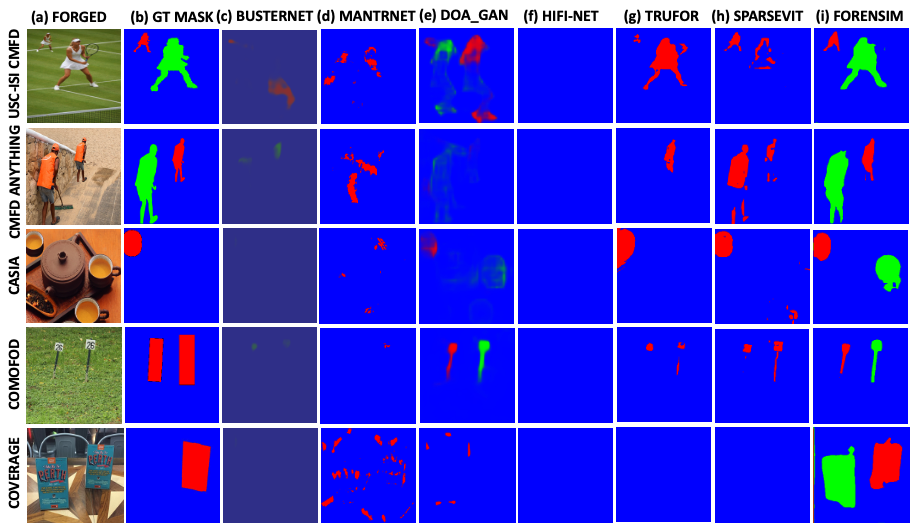}
        \label{fig:qualitative_eval}
    \end{minipage}
\end{figure*}
\endgroup

\vspace{-2px}
\subsection{Motivation and Dataset Gap}
\vspace{-2px}

As summarized in Table 1 of \cite{nandi2023trainfors}, existing CMFD benchmarks include only 941 forged images across four public datasets, none providing three-class RGB masks required to train models like Forensim. USC-ISI CMFD~\cite{wu2018busternet}, based on MS-COCO~\cite{lin2014microsoft} and SUN2012~\cite{xiao2010sun}, contains synthetic images that are often visually obvious and unrealistic. In practice, skilled adversaries can create convincing forgeries difficult to detect—even for humans. Reproducibility is also hindered as USC-ISI CMFD is no longer publicly available, and newer datasets like CatNet~\cite{kwon2021cat} focus only on splicing. These limitations underscore the need for a realistic, large-scale CMFD\_Anything dataset.

\vspace{-2px}
\subsection{CMFD\_Anything Dataset}
\vspace{-2px}

We introduce \texttt{CMFD\_Anything}, a high-resolution benchmark built from SA-1B images and masks~\cite{kirillov2023segment}. Forgeries are created via two strategies: (1) duplicating an object within single-category images, and (2) copying objects from multi-object images to form multiple variants. Each paste is refined with MGMatting~\cite{yu2021mask} and mild clone-stamp–style transforms (translation, small rotation/scale). As existing CMFD datasets lack negative (pristine) samples, we include 100K unaltered images from SA-1B to enable supervised training and false-positive calibration. The final dataset has 200K forged and 100K pristine images, split 8{:}1{:}1 (train/val/test); examples are shown in \hyperref[fig:cmfd_anything]{Fig.~\ref*{fig:cmfd_anything}}.

\noindent\textbf{Copy placement.} We select destinations in two steps: (i) axis-aligned proposals by sliding along the $x$/$y$ axes through the source centroid and along its principal axis, subject to distance/scale/overlap gates; (ii) semantic filtering with OpenCLIP~\cite{cherti2023reproducible} image-only embeddings on local context windows, choosing the arg-max cosine match and requiring $\cos\!\ge\!\tau$ (default $\tau{=}0.30$). We will release per-instance metadata (source/target masks, transform, chosen axis, and similarity) for transparent provenance evaluation.

\noindent\textbf{Quality Assessment.} We evaluate \texttt{CMFD\_Anything} using No-Reference~\cite{ying2020patches, wang2023exploring} (forged-only) and Full-Reference~\cite{wang2004image, zhang2018unreasonable} (forged vs. pristine) metrics. As shown in \hyperref[tab:cmfd_quality_scores]{Tab.~\ref*{tab:cmfd_quality_scores}}, it outperforms CASIA~\cite{dong2013casia} and CoMoFoD~\cite{tralic2013comofod}. To filter low-quality forgeries, we train a binary classifier to reject synthetic samples that are easily identified as fake.



\subsection{Training and Implementation}
Images are resized to \(224 \times 224\) and processed with a DINO-pretrained Vision Transformer backbone. Models are implemented in PyTorch and trained on an NVIDIA RTX A5000 GPU using AdamW (batch size 64). A cyclic learning rate (\num{1e-3} to \num{1e-5}) is used with StepLR decay (0.5 every 10 epochs). Early stopping is applied based on validation loss. Forensim is trained for 100 epochs on 100K randomly sampled images per epoch from the training pool. The model is optimized with Cross-Entropy and InfoNCE losses, with additional trials using Dice and Focal losses to mitigate class imbalance (details in supplementary).

\section{Experimental Evaluation}

\vspace{-2px}
\subsection{Evaluation Datasets and Metrics}
\vspace{-2px}

We evaluate Forensim on three standard copy-move forgery detection (CMFD) benchmarks—USC-ISI CMFD~\cite{wu2018busternet}, CoMoFoD~\cite{tralic2013comofod}, and CASIA CMFD~\cite{dong2013casia}—as well as the test split of our proposed \texttt{CMFD\_Anything} dataset. Additionally, we assess generalization on four tampered image detection and localization (IMDL) datasets: NIST16~\cite{NimbleCh33:online}, Columbia~\cite{ng2009columbia}, Coverage~\cite{wen2016coverage}, and CASIA~\cite{dong2013casia}. Notably, neither CASIA nor CoMoFoD provides source and target masks—only binary annotations (\textcolor{red}{red pixels} in \hyperref[fig:qualitative_eval]{Fig.~\ref*{fig:qualitative_eval}}).

For evaluation, we report both pixel-level and image-level metrics. Pixel-level performance is assessed using precision, recall, F1-score, and AUC across three forgery classes: \textcolor{blue}{pristine}, \textcolor{green}{source}, and \textcolor{red}{target}, derived from three-class RGB masks generated by Forensim and baseline methods. Image-level detection is evaluated using precision, recall, and F1-score. To ensure consistent global evaluation, true positives (TP), false positives (FP), and false negatives (FN) are aggregated across the entire dataset prior to metric computation. Following prior works~\cite{wu2018busternet, wu2019mantra, islam2020doa}, we apply a 200-pixel threshold (for $320 \times 320$ images) to suppress small false positives. 

\hyperref[tab:cmfd_combined]{Tab.~\ref*{tab:cmfd_combined}} and \hyperref[tab:casia_comofod_merged]{Tab.~\ref*{tab:casia_comofod_merged}} present detailed pixel- and image-level results on CMFD datasets. \hyperref[tab:natural_f1]{Tab.~\ref*{tab:natural_f1}} reports pixel-level F1 scores across four IMDL benchmarks, highlighting the generalization of Forensim to diverse manipulation types and real-world forgeries.

\vspace{-5px}
\subsection{Forensim Results and Analysis}
\label{forensim_results_analysis}

For fairness and reproducibility, we restrict comparisons to models with publicly accessible code. All baseline methods were retrained following the protocol in~\cite{kwon2021cat} and evaluated on publicly available benchmark datasets. We compare Forensim against models trained on binary masks across both CMFD-specific and IMDL datasets, validating the effectiveness of our three-class training formulation. As shown in \hyperref[tab:cmfd_combined]{Tab.~\ref*{tab:cmfd_combined}}, \hyperref[tab:casia_comofod_merged]{Tab.~\ref*{tab:casia_comofod_merged}}, and \hyperref[tab:natural_f1]{Tab.~\ref*{tab:natural_f1}}, Forensim consistently achieves the best F1 and AUC scores for both pixel-level localization and image-level detection across all benchmarks.

\begin{figure}[t]
  \centering
  \begin{subfigure}{0.53\linewidth}
    \centering
    \includegraphics[width=\linewidth]{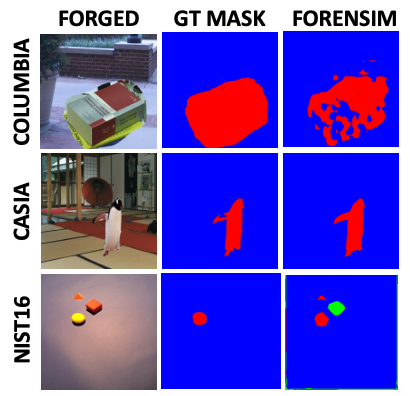}
    \caption{Benchmark IMDL Localization}
    \label{fig:forensim_imdl}
  \end{subfigure}\hfill
  \begin{subfigure}{0.47\linewidth}
    \centering
    \includegraphics[width=\linewidth]{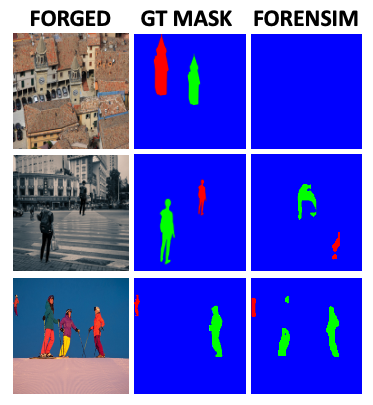}
    \caption{Misdetection and False Alarms}
    \label{fig:misdetection_false_alarm}
  \end{subfigure}
  \caption{(a) Forensim IMDL; (b) Forensim Limitations}
  \label{fig:forensim_combined}
\end{figure}

Forensim independently models duplication similarity (\texttt{Sim\_Map}) and manipulation structure (\texttt{MSSA\_Map}), subsequently refined through non-local aggregation. It outputs three-class source-target masks—outperforming SOTA IMDL baselines in qualitative evaluation (\hyperref[fig:qualitative_eval]{Fig.~\ref*{fig:qualitative_eval}}) and outputs binary forgery masks for splicing, when no in-image source exists (\hyperref[fig:forensim_imdl]{Fig.~\ref{fig:forensim_imdl}}). Despite its strong performance, Forensim remains efficient—36.7M params and 28.7 GFLOPs at $512\times512$ resolution, while outperforming larger baselines like TruFor~\cite{guillaro2023trufor}, SparseViT~\cite{su2025can} (\hyperref[tab:complexity]{Tab.~\ref*{tab:complexity}}).

\vspace{-5px}
\subsection{Ablation Analysis and Robustness}
\vspace{-2px}

We perform ablations on three Forensim components: \texttt{Sim\_Attn}, \texttt{MSSA}, and the detection branch (\hyperref[tab:cmfd_combined]{Tab.~\ref*{tab:cmfd_combined}}, \hyperref[tab:casia_comofod_merged]{Tab.~\ref*{tab:casia_comofod_merged}}). To further understand key design choices, we analyze normalization strategies and pre-processing conditions. \hyperref[tab:affinity_norm_ablation]{Tab.~\ref*{tab:affinity_norm_ablation}} shows pixel-level F1 scores across three variants of the \texttt{Sim\_Attn} normalization. The BiSoftmax configuration, which applies full normalization along both row and column dimensions, consistently outperforms the raw and Row-only variants across datasets. This demonstrates the importance of bi-directional affinity normalization for capturing discriminative similarity cues in IMDL.

\hyperref[tab:cmfd_ablation_extended]{Tab.~\ref*{tab:cmfd_ablation_extended}} presents additional ablations on the \texttt{CMFD\_Anything} benchmark by varying mask generation strategies. Removing the MGMatting post-processing, replacing three-class masks with binary masks, or introducing noise into ground truth annotations each lead to degraded performance across all input resolutions. In contrast, the full \texttt{CMFD\_Anything} pipeline achieves the best F1 scores, particularly at $512 \times 512$, highlighting the value of high-fidelity soft masks and precise annotations for robust model training.

\begin{figure}[t]
    \centering
    \includegraphics[width=\columnwidth]{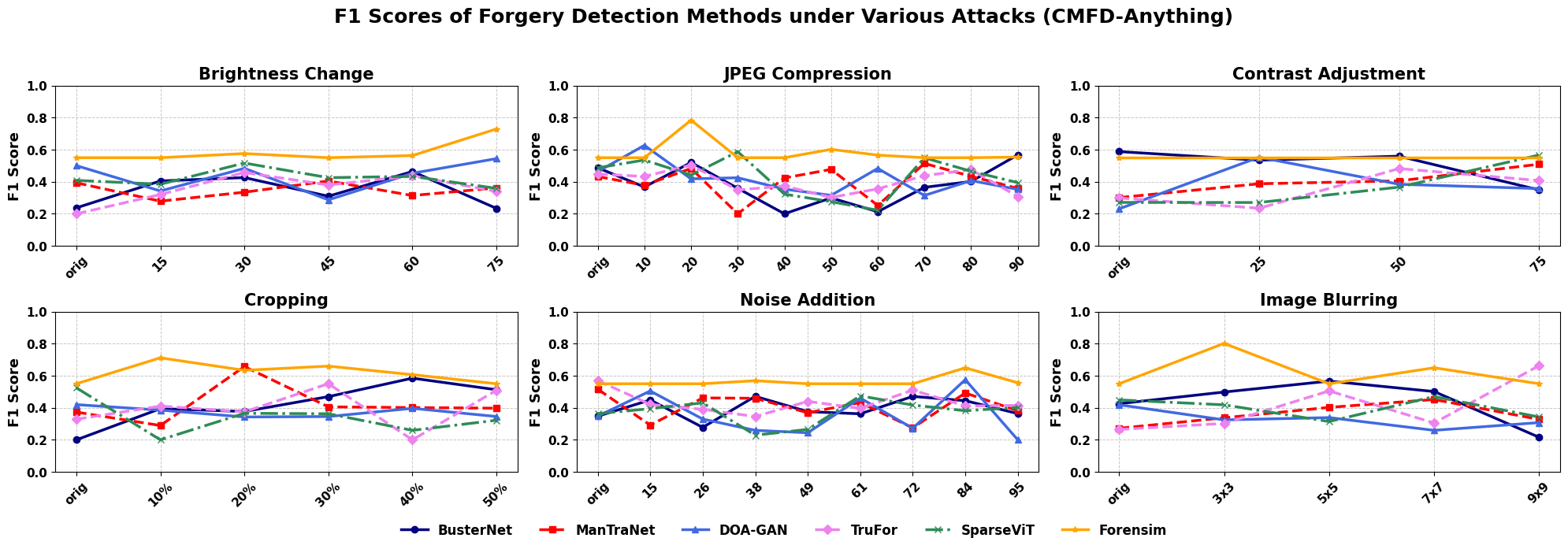} 
    \caption{Robustness under attacks on CMFD\_Anything.}
    \label{fig:forensim_robustness}
\end{figure}

To assess robustness, we test Forensim under six perturbations—brightness, JPEG compression, contrast, cropping, noise, and blur—following~\cite{wu2018busternet,wu2019mantra}, plus four social-media perturbations (supplementary, Table~4) following~\cite{wu2022robust}. As shown in \hyperref[fig:forensim_robustness]{Fig.~\ref*{fig:forensim_robustness}}, Forensim consistently outperforms state of the art across all perturbations.

These results underscore Forensim’s design: \texttt{Sim\_Attn} captures global duplication, \texttt{MSSA} local artifacts, and multi-scale state-space attention enhances robustness; with three-class supervision, Forensim demonstrates strong generalization across forgery types and image-level distortions.


\begin{table}[t]
    \centering
    \caption{Pixel-level F1 localization on benchmark IMDL datasets using best threshold and a fixed threshold of 0.5 following~\cite{guillaro2023trufor}.}
    \vspace{-2mm}
    \scriptsize
    \setlength{\tabcolsep}{2.5pt}
    \begin{tabular}{l|cc|cc|cc|cc|cc}
        \toprule
        \textbf{Method} &
        \multicolumn{2}{c|}{\textbf{NIST}~\cite{NimbleCh33:online}} &
        \multicolumn{2}{c|}{\textbf{COL}~\cite{ng2009columbia}} &
        \multicolumn{2}{c|}{\textbf{COV}~\cite{wen2016coverage}} &
        \multicolumn{2}{c|}{\textbf{CASIA}~\cite{dong2013casia}} &
        \multicolumn{2}{c}{\textbf{Avg.}} \\
        \cmidrule{2-11}
        & Best & Fixed & Best & Fixed & Best & Fixed & Best & Fixed & Best & Fixed \\
        \midrule
        ManTra~\cite{wu2019mantra}      & 21.9 & 19.3 & 47.5 & 46.2 & 21.1 & 19.6 & 38.2 & 32.7 & 32.2 & 29.5 \\
        TruFor~\cite{guillaro2023trufor} & 35.6 & 34.8 & 89.4 & 88.5 & 47.3 & 45.7 & 83.5 & 81.8 & 63.9 & 62.7 \\
        SparseViT~\cite{su2025can}      & \underline{39.4} & \underline{38.4} & \textbf{95.9} & \textbf{95.9} & \underline{52.5} & \underline{51.3} & \textbf{84.2} & \textbf{82.7} & \underline{68.0} & \underline{67.1} \\
        \textbf{Forensim (Ours)}        & \textbf{40.2} & \textbf{39.1} & \underline{94.2} & \underline{93.8} & \textbf{55.7} & \textbf{54.6} & \underline{83.4} & \underline{81.8} & \textbf{68.4} & \textbf{67.3} \\
        \bottomrule
    \end{tabular}
    \vspace{-2mm}
    \label{tab:natural_f1}
\end{table}



\vspace{-2px}
\subsection{Failure Cases and Future Directions}
\vspace{-2px}

While \textit{Forensim} is robust overall, we observe three recurring failure modes: (i) background-matching splices where low contrast suppresses \(\mathrm{MSSA}_{\text{map}}\) (row~1, Fig.~\ref{fig:misdetection_false_alarm}); (ii) repetitive structures causing many-to-many matches in \(\mathrm{Aff}_{\text{final}}\) (row~3); and (iii) rare errors on low-color/monochrome images. Remedies include a lightweight frequency/boundary head and photometric/codec augmentations for (i); uniqueness constraints (entropy-based or non-max suppression) and hard-negative mining for (ii); and grayscale-focused augmentation with contrast-limited adaptive histogram equalization plus a luminance similarity head for (iii). We will broaden 2D RoPE and extend to video IMDL with spatio-temporal consistency.

\begin{table}[t]
    \centering
    \scriptsize
    \caption{Pixel-level F1 across normalization types in Sim\_Attn.}
    \vspace{-2mm}
    \setlength{\tabcolsep}{3.5pt}
    \begin{tabular}{l|c|c|c|c}
        \toprule
        \textbf{Variant} & \textbf{Norm.} & \textbf{CASIA CMFD~\cite{dong2013casia}} & \textbf{CoMoFoD~\cite{tralic2013comofod}} & \textbf{CMFD\_Anything} \\
        \midrule
        None         & Raw     & 31.6  & 29.2 & 33.8 \\
        Row-only     & Partial & \underline{44.8}  & \underline{36.7} & \underline{54.3} \\
        \textbf{BiSoftmax} & Full    & \textbf{58.1}  & \textbf{47.8} & \textbf{62.4} \\
        \bottomrule
    \end{tabular}
    \vspace{-2mm}
    \label{tab:affinity_norm_ablation}
\end{table}

\begin{table}[t]
    \centering
    \scriptsize
    \caption{Ablation studies on CMFD\_Anything using Forensim.}
    \vspace{-2mm}
    \setlength{\tabcolsep}{3.5pt}
    \begin{tabular}{l|c|c|c|c}
        \toprule
        \textbf{Ablation Condition} & \textbf{$224 \times 224$} & \textbf{$256 \times 256$} & \textbf{$320 \times 320$} & \textbf{$512 \times 512$} \\
        \midrule
        No MGMatting        & 56.6 & 57.7 & 58.5 & 61.2 \\
        Binary Mask         & 53.9 & 55.2 & 56.3 & 59.4 \\
        Noisy Groundtruth            & 53.4 & 54.6 & 55.5 & 58.4 \\
        \textbf{CMFD\_Anything} & \textbf{59.6} & \textbf{60.8} & \textbf{62.1} & \textbf{62.4} \\
        \bottomrule
    \end{tabular}
    \vspace{-2mm}
    \label{tab:cmfd_ablation_extended}
\end{table}


\begin{table}[t]
    \centering
    \scriptsize
    \caption{Model complexity w.r.to Resolution, Parameters, FLOPs.}
    \vspace{-2mm}
    \setlength{\tabcolsep}{3.5pt}
    \begin{tabular}{l|c|c|c|c}
        \toprule
        \textbf{Method} & \textbf{Backbone} & \textbf{Input Size} & \textbf{Params} & \textbf{FLOPs} \\
        \midrule
        BusterNet~\cite{wu2018busternet} & VGG    & $256 \times 256$ & 15.5M  & 45.7G \\
        ManTraNet~\cite{wu2019mantra}    & VGG    & $256 \times 256$ & 3.9M   & 274.0G \\
        DOA\_GAN~\cite{islam2020doa}     & VGG    & $256 \times 256$ & 26.9M  & 46.7G \\
        TruFor~\cite{guillaro2023trufor} & ViT   & $512 \times 512$ & 68.7M  & 236.5G \\
        SparseViT~\cite{su2025can}       & ViT   & $512 \times 512$ & 50.3M  & 46.2G \\
        \textbf{Forensim (Ours)}         & SSM    & $512 \times 512$ & 36.7M  & 28.7G \\
        \bottomrule
    \end{tabular}
    \vspace{-2mm}
    \label{tab:complexity}
\end{table}



\vspace{-2px}
\section{Conclusion}
\label{sec:conclusion}
\vspace{-2px}

We presented Forensim, a unified framework for IMDL that accurately localizes both source and forged regions using three-class supervision. We propose similarity and manipulation attention modules within a state-space framework, enabling efficient modeling of long-range dependencies with linear complexity. Forensim addresses key limitations of existing CMFD and splicing models, which typically rely on binary masks and artifact-based cues. To support robust training and evaluation, we introduced \texttt{CMFD\_Anything}, a high-resolution dataset containing realistic copy-move forgeries. Extensive experiments across multiple benchmarks demonstrate that Forensim achieves state-of-the-art performance and generalizes effectively across manipulation types and attack scenarios, establishing it as a unified and interpretable solution for image forensics.

{
    \small
    \bibliographystyle{ieeenat_fullname}
    \bibliography{main}
}


\end{document}